\def\year{2020}\relax
\documentclass[letterpaper]{article} 
\usepackage{aaai20}  
\usepackage{times}  
\usepackage{helvet} 
\usepackage{courier}  
\usepackage[hyphens]{url}  
\usepackage{graphicx} 
\usepackage{graphicx}  
\nocopyright
\usepackage{subcaption}

\title{PulseSatellite:  A tool using human-AI feedback loops for satellite image analysis in humanitarian contexts}
\author{Tomaz Logar,\textsuperscript{\rm 1} Joseph Bullock,\textsuperscript{\rm 1}\textsuperscript{,\rm 2}  Edoardo Nemni,\textsuperscript{\rm 3} Lars Bromley,\textsuperscript{\rm 3} John A. Quinn,\textsuperscript{\rm 1} Miguel Luengo-Oroz\textsuperscript{\rm 1}\\ 
\textsuperscript{\rm 1}United Nations Global Pulse
\textsuperscript{\rm 2}Institute for Data Science, Durham University
\textsuperscript{\rm 3}UNOSAT\\
tomaz@unglobalpulse.org, miguel@unglobalpulse.org
}

\begin{document}

\maketitle

\begin{abstract}
Humanitarian response to natural disasters and conflicts can be assisted by satellite image analysis. In a humanitarian context, very specific satellite image analysis tasks must be done accurately and in a timely manner to provide operational support. We present PulseSatellite, a collaborative satellite image analysis tool which leverages neural network models that can be retrained on-the fly and adapted to specific humanitarian contexts and geographies. We present two case studies, in mapping shelters and floods respectively, that illustrate the capabilities of PulseSatellite. 
\end{abstract}

\section{Introduction}

When dealing with conflict and humanitarian crises, accurate and timely satellite image analysis is key to supporting critical operations on the ground. Use cases include monitoring population displacement, settlement mapping, damage assessment, fire detection associated with human rights violations, damage to transportation networks, floods assessment or identifying direct impact of earthquakes, volcanoes, cyclones and landslides \cite{lang2015humanitarian}. In these contexts, automated processes providing vital information in decision making must be carefully validated and tuned to maximal performance since e.g. a false positive may be detrimental to human life. While there exist several general purpose satellite image analysis tools, few are designed and optimized for humanitarian use cases. PulseSatellite is a tool to analyze satellite imagery assisted by neural networks that seeks to incorporate humans-in-the-loop at different stages in the model inference process to enable optimal results and expert validation in humanitarian contexts. The conceptual framework used to implement and deploy PulseSatellite has been previously presented in \cite{camp_paper}.

\section{PulseSatellite Functionalities}

\begin{figure}[h]
\centering
\includegraphics[width=0.99\linewidth]{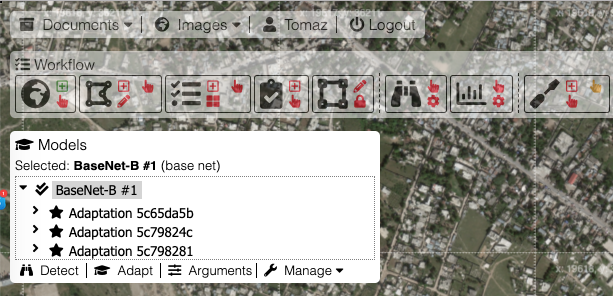}
\caption{Detail of PulseSatellite showing a panel that allows the selection of a trained network from a hierarchy of models adapted to different use cases.} 
\label{camp_oscreenshot}
\end{figure}

\begin{figure}[h]
\centering
\includegraphics[width=0.92\linewidth]{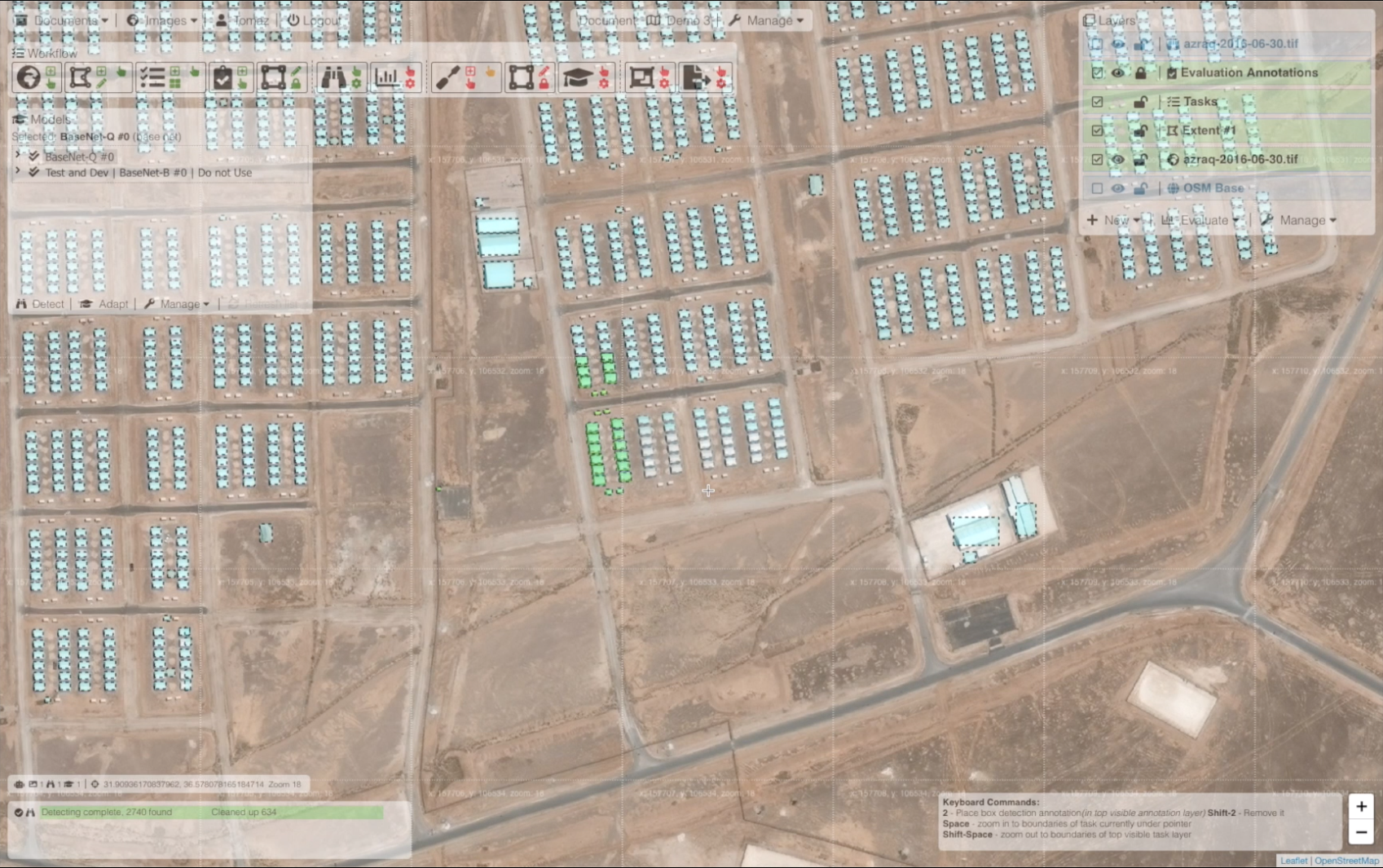}
\caption{PulseSatellite mapping a refugee camp in Jordan. Model detected buildings are highlighted in blue, with analyst evaluation structures in green.}
\label{camp_screenshot}
\end{figure}

The tool enables the performance of such specific humanitarian tasks as identifying and counting structures in refugee settlements and rapid mapping flooded areas after a natural disaster. These tasks can be repetitive in nature, and hence lend themselves to support with AI tools; however, each specific analysis scenario can have distinct features (such as a certain type of shelter construction) which are not all feasible to handle with a single pre-trained system. Therefore, the PulseSatellite system allows users to fine-tune a hierarchy of models as shown in Figure \ref{camp_oscreenshot}, thus including expert knowledge through a human-in-the-loop system. For example, the generic shelter detection model may be fine tuned progressively to a model specialized for desert terrain, then to another which is specific to a different scenario (such as the camp in Figure \ref{camp_screenshot}). Any of these models can then be used as starting points for new scenarios. On-the-fly evaluation helps the analyst to determine whether the assessment quality is good enough for release, in which case detection results can be exported. These steps can all be carried out via the web front end and with multiple remote collaborators.


\section{PulseSatellite Machine Learning Framework}

PulseSatellite is designed to host a variety of specialised models, fine-tuned for different functionalities. Given that each scenario may require different outputs, the tool must be dynamic such that new models can be trained in real-time (as users provide feedback on detection results), and easily adaptable to new situations. In this section we explore a series of case studies demonstrating a cross-section of humanitarian contexts PulseSatellite is being used to tackle.

\subsection{System Architecture}

PulseSatellite is a distributed cloud application, currently running on Amazon Web Services (AWS). The web-based user interface (UI) is written in JavaScript+HTML, and the web server, which communicates with the UI via HTTP and WS protocols, is written in NodeJS. The database is managed using PostgreSQL with PostGIS extension and is currently also used as a message streaming service for Pub/Sub messaging between the web, GPU and tiling servers. All backend files are generated and stored on a Amazon Elastic File System (EFS). Multiple GPU servers can run ML processes across a range of frameworks including PyTorch, TensorFlow and Keras. GIS processes, including the tiling servers, use OSGeo libraries and programs, most notably GDAL. All used software is open source.

\subsection{Case Study: Mapping Refugee Settlements}

Counting and classifying structures in a refugee settlement is a common analysis task for humanitarian agencies. In practice, this is currently manually done by human expert analysts using satellite imagery. A single settlement may have tens of thousands of structures, and identifying each of them can take several days. For camp mapping, we have a Mask R-CNN model \cite{mask} trained on images from 12 settlements, where the image of each has been split into 300x300 pixel tiles and annotated by human experts. Once the model has been run on an unseen camp, the analyst can inspect the result in the tool and correct the outputs on a subset of tiles. An adaptation stage can then be performed to fine-tune the model to the unseen image and increase performance. In recent independent tests camp completion rates increased from 77.3\% to 94.7\% after adaptation, with a final user accuracy of 94.4\% -in line with humanitarian performance requirements described in \cite{camp_paper}.

\subsection{Case Study: Rapid Flood Mapping}


Rapid analysis of regions affected by natural disasters is essential for response planning.  Current methods for rapid flood mapping require an analyst to perform a series of manual pixel intensity thresholding and image processing operations, taking several hours to be completed. The PulseSatellite tool uses a U-Net model \cite{unet} to produce a flood map in the order of minutes. The model has been trained and tested on images from Bangladesh and Somalia achieving overall accuracy scores of over 90\%, in line with current manual methods. 

\section{Impact and Discussion}
The PulseSatellite camp mapping function has been piloted in operational conditions in multiple settlements across Africa and the Middle East. Results revealed that while high levels of accuracy are possible with machine learning methods, there is considerable variation in the characteristics of new imagery in real life scenarios requiring analysis workflows that allow the incorporation of automatic methods in a progressive manner. Using AI assistants as an augmentation of human analysts is a reasonable strategy to transition from current fully manual operational pipelines to ones which are both more efficient and have the necessary levels of quality control. PulseSatellite is expected to allow human analysts to drastically reduce the time required to have high accuracy detection in particular use cases. Future specific AI modules and front-end functionalities to be added in PulseSatellite include change detection for cultural heritage preservation and damage assessment. PulseSatellite is available to international and humanitarian organizations. 

\section{Acknowledgments}
United Nations Global Pulse is supported by the Governments of Netherlands, Sweden and Germany and the William and Flora Hewlett Foundation. The United Nations Institute for Training and Research’s (UNITAR) Operational Satellite Applications Programme (UNOSAT) is supported by the Norwegian Agency for Development Cooperation. JB also is supported by UK Science and Technology Facilities Council (STFC) grant number ST/P006744/1.

\bibliographystyle{aaai}
\bibliography{aaai}

\begin{thebibliography}{}

\bibitem[\protect\citeauthoryear{He \bgroup et al\mbox.\egroup }{2017}]{mask}
He, K.; Gkioxari, G.; Doll{\'a}r, P.; and Girshick, R.~B.
\newblock 2017.
\newblock Mask r-cnn.
\newblock {\em 2017 IEEE International Conference on Computer Vision (ICCV)}
  2980--2988.

\bibitem[\protect\citeauthoryear{Lang \bgroup et al\mbox.\egroup
  }{2015}]{lang2015humanitarian}
Lang, S.; F{\"u}reder, P.; Kranz, O.; Card, B.; Roberts, S.; and Papp, A.
\newblock 2015.
\newblock Humanitarian emergencies: causes, traits and impacts as observed by
  remote sensing.
\newblock In {\em Remote Sensing of Water Resources, Disasters, and Urban
  Studies}. CRC Press.
\newblock  483--512.

\bibitem[\protect\citeauthoryear{Quinn \bgroup et al\mbox.\egroup
  }{2018}]{camp_paper}
Quinn, J.~A.; Nyhan, M.~M.; Navarro, C.; Coluccia, D.; Bromley, L.; and
  Luengo-Oroz, M.
\newblock 2018.
\newblock Humanitarian applications of machine learning with remote-sensing
  data: review and case study in refugee settlement mapping.
\newblock {\em Philosophical Transactions of the Royal Society A: Mathematical,
  Physical and Engineering Sciences.} 376.

\bibitem[\protect\citeauthoryear{Ronneberger, Fischer, and Brox}{2015}]{unet}
Ronneberger, O.; Fischer, P.; and Brox, T.
\newblock 2015.
\newblock U-net: Convolutional networks for biomedical image segmentation.
\newblock {\em Medical Image Computing and Computer-Assisted Intervention
  (MICCAI)} 9351:234--241.

\end{thebibliography}

\end{document}